\documentclass[letterpaper,10pt,conference]{ieeeconf}
\IEEEoverridecommandlockouts
\usepackage{fontspec}
\usepackage{amsmath,amssymb,booktabs,graphicx}
\usepackage{stfloats}
\usepackage{cite}
\usepackage[hidelinks]{hyperref}
\hypersetup{pdfauthor={Qianxing Li},pdftitle={AnyBodyCodec: A Shared Native Interface for Body-Model Conversion, Extension, and Humanoid Control}}
\newcommand{\method}{AnyBodyCodec}
\title{\LARGE\bf AnyBodyCodec: A Shared Native Interface for Body-Model\\
Conversion, Extension, and Humanoid Control}
\author{Qianxing Li%
\thanks{Qianxing Li is affiliated with Delta Intelligence and China University of Geosciences (Beijing).
Email: \texttt{\href{mailto:leeepic19@gmail.com}{leeepic19@gmail.com}}.}%
}
\begin{document}
\maketitle
\thispagestyle{empty}\pagestyle{empty}
\begin{figure*}[!b]
\centering
\includegraphics[width=\textwidth]{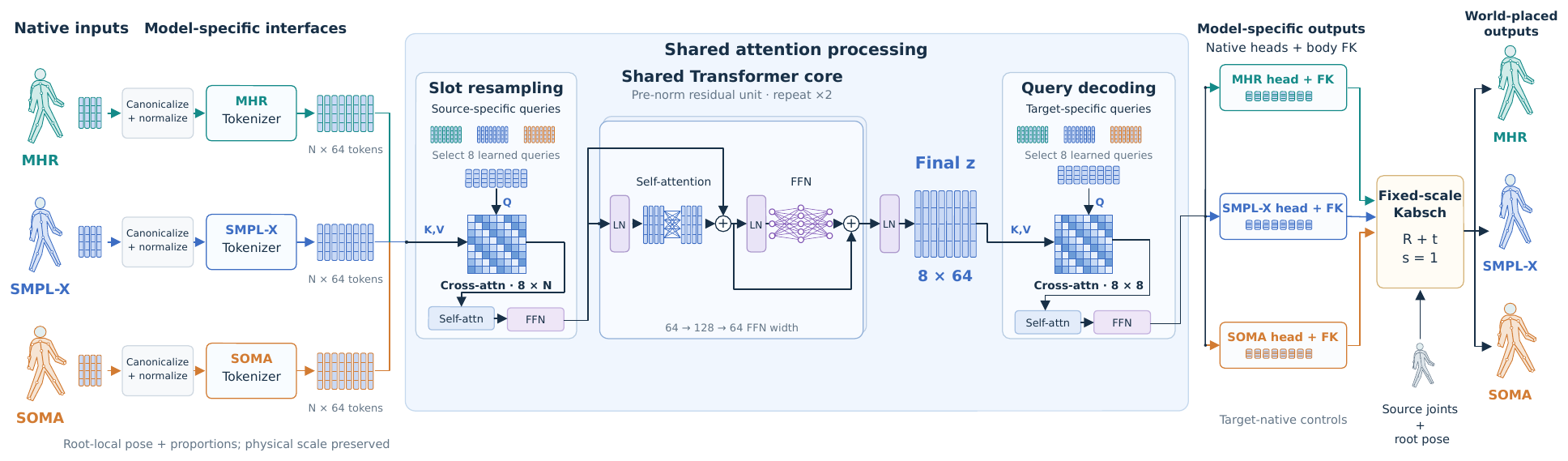}
\caption{Native conversion through shared slots. Model-specific tokenizers and queries
encode root-local controls. A two-layer Transformer and final LayerNorm produce
$z\in\mathbb R^{8\times64}$. Target queries and native heads decode body-layer controls;
input-guided fixed-scale Kabsch restores world rotation and translation.}
\label{fig:method-overview}
\end{figure*}

\begin{abstract}
Parametric body models expose incompatible pose and shape controls, tying humanoid
policies to particular input formats. AnyBodyCodec learns a shared representation
through native conversion, then reuses it for new body models and control with its
core frozen. Encoders map MHR, SMPL-X, and SOMA to aligned slots; native decoders
recover target controls. The 1.090M-parameter converter covers nine routes and reaches
16.801 mm input-reference MPJPE across six cross routes: the 2.887M Transformer bank
reaches 16.390 mm and its parameter-matched version 18.866 mm. Three measured routes
take 7.43--7.67 ms on an RTX 4090. Adding ANNY shows that comparable conversion accuracy can accompany
different action consistency under one frozen controller. Under a matched PPO budget,
aligned slots support 60.78\% full-action success across three formats. Separate
160-hour training reaches 66.10\% source-weighted validation success, and the controller
is deployed qualitatively on a physical G1 through three visual front ends. Together,
these results establish reuse through both native geometric fidelity and downstream
policy compatibility.
\end{abstract}
\section{Introduction}
\label{sec:introduction}
A humanoid controller should be able to follow the same human motion when its
perception system changes the body model used to describe it. Yet parametric body
models expose incompatible pose and shape controls, tying conversion tools and policy
inputs to particular formats. We study whether native body-model conversion can learn
a shared motion representation that remains useful across changes in body-model
interfaces and downstream tasks.

Such reuse requires geometric fidelity and consistent interpretation. The representation
must retain enough information to recover target-native controls, while allowing a
fixed policy to interpret corresponding motions across formats. Geometric agreement
alone does not establish this compatibility: native decoders may recover similar poses
from features that elicit different actions from the same controller.

We introduce \method{} as a shared native interface for MHR, SMPL-X, and SOMA
(Fig.~\ref{fig:method-overview}). Model-specific encoders map paired native fits to
aligned slots that constitute the shared representation; target decoders recover native pose
and shape controls. We then freeze the conversion-trained network and learn new
body-model interfaces or a control adapter around it. Our contribution is this reusable
representation and its evaluation across native conversion, frozen extension, and
humanoid control.

The experiments establish this reuse in three steps. Compact conversion approaches
the accuracy of a Transformer bank with 2.65 times as many parameters and improves
on a parameter-matched bank (Table~\ref{tab:conversion}). Attaching ANNY tests whether
new interfaces can learn around the fixed representation. Similar conversion errors
but different frozen-policy action deviations reveal why downstream compatibility
needs its own evaluation (Fig.~\ref{fig:anny-policy-consistency}). Finally, matched-budget
PPO evaluates shared-policy learning across three formats (Table~\ref{tab:control});
separate 160-hour training tests larger-corpus learning (Table~\ref{tab:scaling}).
Qualitative G1 deployment through three visual front ends demonstrates reuse across
perception formats.

\section{Related Work}
\textbf{Native conversion and retargeting.}
MHR and SMPL-X expose native articulation and proportion controls
\cite{ferguson2025mhr,pavlakos2019smplx}. SOMA addresses unification and compatibility
among parametric body models through a common body-model representation
\cite{saito2026soma}. Neural Kinematic Networks combine kinematics and cycle
learning~\cite{villegas2018nkn}, while skeleton-aware networks share structure across
homeomorphic skeletons~\cite{aberman2020skeletonaware}. AnyBodyCodec retains dedicated
encoders and decoders for each model's native controls and learns a shared latent
interface through conversion. We further examine whether that interface preserves a
frozen controller's interpretation of motion across native inputs.

\textbf{Shared motion representations.}
SAME and HuMoT learn skeleton-agnostic embeddings for motion retargeting
\cite{lee2023same,mourot2023humot}; AnyTop generates motion across topologies
\cite{gat2025anytop}. These works study motion transfer or generation across skeleton
structures. Perceiver IO supplies the learned-query mechanism used for our native
encoding and decoding~\cite{jaegle2021perceiverio}. Our focus is on learning shared
features that recover heterogeneous models' native pose and shape controls, then
testing whether the fixed representation supports new native interfaces and humanoid
control.

\textbf{Humanoid motion control.}
SONIC studies high-performance humanoid motion tracking through a learned motion-token
interface for whole-body control~\cite{luo2026sonic}. We adopt a SONIC-style PPO
architecture to evaluate a representation learned before policy training through
heterogeneous native conversion. The body representation stays frozen while the
control networks learn from scratch. Our question is whether new native input
interfaces preserve how the same policy interprets motion, linking representation
reuse to downstream control compatibility.

\section{A Shared Native Body Representation}
\label{sec:method}
\subsection{Native Encoding and Decoding}
For model $m\in\{\mathrm{MHR},\mathrm{SMPL\mbox{-}X},\mathrm{SOMA}\}$,
native controls $x^m$ specify pose, proportions, and root placement. Canonicalization
$C_m$ sets root rotation to identity and centers the forwarded pelvis, retaining bone
lengths. Fixed training statistics standardize each parameter group; source values are
clipped to $[-5,5]$. The conversion route is
\begin{equation}
\begin{aligned}
 z^s&=E_s(N_s(C_s(x^s)))\in\mathbb R^{8\times64},\\
 \hat x^t&=N_t^{-1}(D_t(z^s)).
\end{aligned}
 \label{eq:conversion}
\end{equation}
The slots $z$ form the shared latent representation. Native controls are body-model
parameters; joint coordinates are obtained from the body layer.
The target body layer forwards $\hat x^t$ with a canonical root. Target supervision
uses unclipped normalized controls, and inverse standardization restores native units.

Each scalar forms a token with learned value scaling and parameter-index, group,
and model embeddings. Validity masks identify padding. MHR additionally uses two gated
parameter-token self-attention layers. Eight source-specific queries resample the
tokens; two shared pre-norm Transformer layers and a final LayerNorm produce $z$.
Eight target-specific queries attend to $z$, and the flattened output feeds a
LayerNorm/linear head for each native group. The attention width is 64 with four heads,
one resampler block, one query-decoder block, FFN expansion two, and zero dropout.
The resampler and query decoder each include cross-attention, query self-attention,
and an FFN. Slots are learned latent positions, not anatomical joints.

\subsection{Paired Correspondence and Geometry}
Training uses paired native controls and root-local target joints. Geometry loss
is smooth-$L_1$ on the 19-joint residual divided by 0.1 m, weighted by fitting
confidence and validity. Native auxiliary losses supervise shape/scale and local pose:
masked smooth-$L_1$ for scalar controls and geodesic error for non-root rotations.
Body layers stay fixed, with gradients propagated through their predicted controls.

Let $v_i^m$ be the $L_2$-normalized flattening of $z_i^m$. Same-frame cosine loss
$\mathcal L_{\rm cos}$ averages $1-\langle v_i^m,v_i^n\rangle$ across model pairs.
Ordered-slot matching uses bidirectional cross-entropy over the eight corresponding
slot indices, with normalized slot dot products at temperature 0.1. Squared off-diagonal
slot cosine defines diversity $\mathcal L_{\rm div}$. The objective is
\begin{equation}
\begin{split}
\mathcal L={}&\mathcal L_{\rm geo}+0.1\mathcal L_{\rm shape}
 +0.1\mathcal L_{\rm local}+10^{-3}\mathcal L_{\rm div}\\
 &+\gamma(q)(0.0125\mathcal L_{\rm cos}+0.00625\mathcal L_{\rm match}).
\end{split}
\label{eq:loss}
\end{equation}
Here $q$ is the training update. The paired weights ramp up over 20k updates, stay full until 50k, and cosine-decay
to 10\% by 190k, remaining there through 945k. Thus correspondence is supervised
directly in the decoder-input tensor while native reconstruction continues.

World placement is separate from the network. Proper Kabsch~\cite{kabsch1976rotation}
fits the predicted target common joints to observed source joints with $R\in SO(3)$,
translation $t$, and scale fixed to one. The input supplies all registration
correspondences. The network predicts root-local articulation and proportions; the
rigid solve restores world placement without changing their scale.

\subsection{Adding a Native Interface}
\label{sec:extension}
\begin{figure}[t]
\centering
\includegraphics[width=\columnwidth]{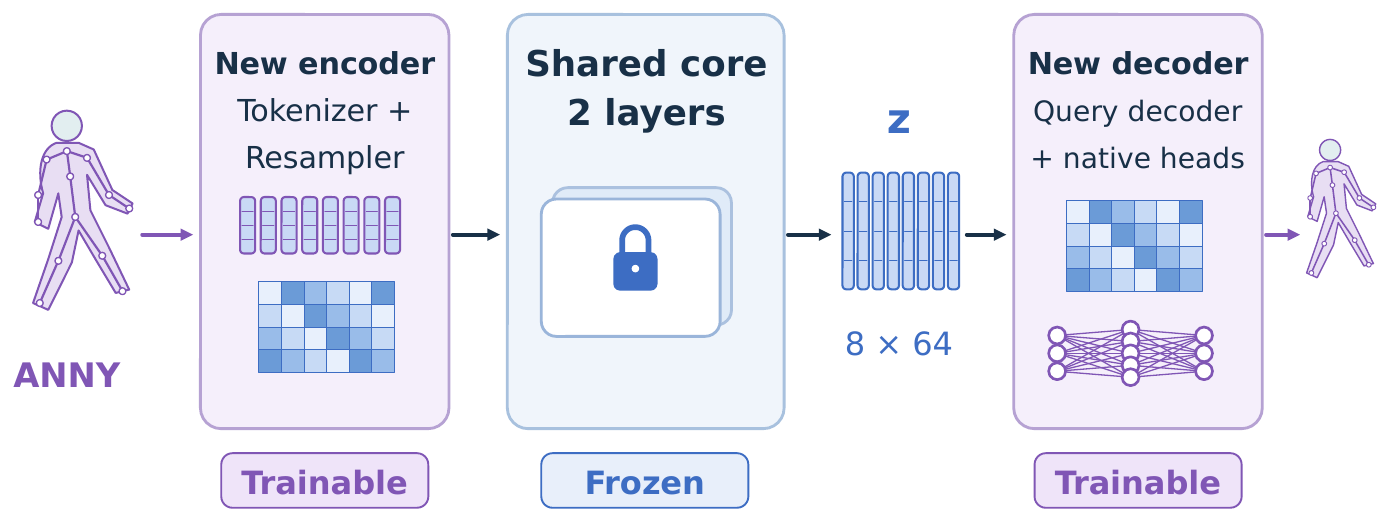}
\caption{ANNY extension. New tokenizer/resampler and query-decoder/native-head
interfaces train around the frozen Transformer. All original interfaces remain frozen.}
\label{fig:anny-extension}
\end{figure}
To attach ANNY, we freeze the complete parent and train its new tokenizer, input
resampler, output-query decoder, and native heads (Fig.~\ref{fig:anny-extension}).
Attention blocks inherit corresponding parent weights; query banks initialize from
the mean of the existing banks. Training uses geometry, shape, local-pose losses and
slot diversity on ANNY-source routes. The new interfaces add 523,961 trainable
parameters to the 1,090,081-parameter parent. Existing routes retain the same frozen
computation.

\section{Native Conversion and Extension}
\label{sec:conversion-results}
\subsection{Data, Training, and Evaluation}
\begin{figure}[!t]
\centering
\includegraphics[width=\columnwidth]{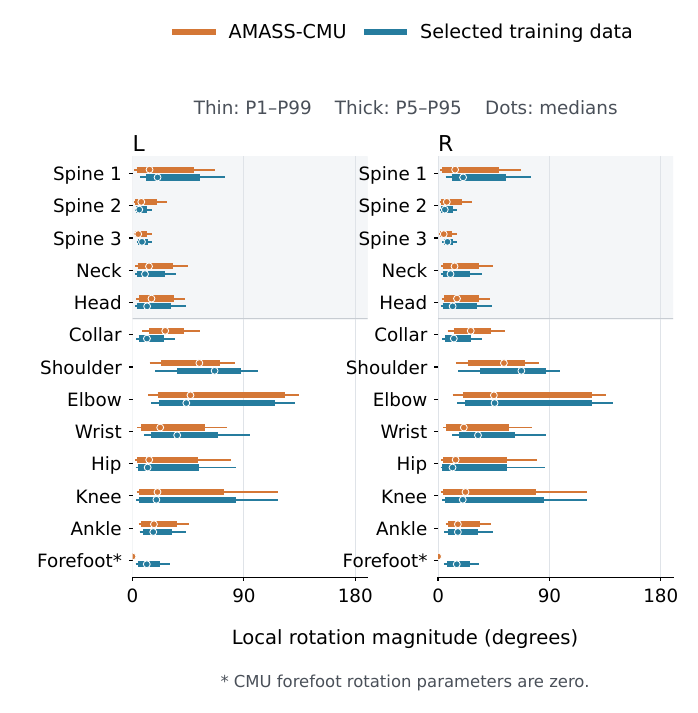}
\caption{Local rotation magnitudes for AMASS-CMU (SMPL-H) and selected training data
(SMPL-X). L/R columns show bilateral joints; shared trunk joints are repeated in both.
Thin/thick intervals denote P1--P99/P5--P95; dots show medians. AMASS frames are duration-
weighted and training pools by size. CMU forefoot parameters are zero (*).}
\label{fig:joint-rotation-coverage}
\end{figure}

Same-frame fits derive from BONES-SEED~\cite{bonesseed2026}: 3,137,326 accepted
records from 1,930 motions and 179 identities, including 1,536,668 mirrors assigned
to their source identity and split. Data selection is guided by the joint-rotation
and identity-parameter distributions observed in real captures, aiming to cover
the parameter ranges encountered in practice. Rows provide native controls, masks, and fitting
confidences for MHR/SMPL-X/SOMA; ANNY fits use the same row indices. The validation set
contains 9,009 frames: 3,024 blocked, 3,500 new-motion, and 2,485 new-identity frames.
All three cohorts contain actor-motion pairs absent from training; the two actors in the
new-identity cohort are also absent from training. Means average frames within each cohort,
then weight routes and cohorts equally. Fig.~\ref{fig:joint-rotation-coverage} compares
the combined original and added training data with AMASS-CMU. For bilateral shoulders, elbows, hips, knees,
and ankles, the training P1--P99 intervals contain 95.2--98.9\% of AMASS-CMU capture time.

The primary metric preserves observed input geometry. SMPL-X-topology joint
regression and a frozen MHR regression backend provide one common 19-joint convention;
ANNY evaluation uses the same regressor on its SMPL-X-topology mesh. After unit-scale
rigid registration to source joints $Q^s$, input-reference MPJPE is
\begin{equation}
 e_{s\to t}=\frac{1000}{19}\sum_{j=1}^{19}
 \|R^*\hat Q_j^t+t^*-Q_j^s\|_2\quad\text{(mm)}.
 \label{eq:metric}
\end{equation}
Training targets are paired fits; this metric measures agreement with the observed
input rather than independent measured human ground truth.

All conversion systems receive 945k updates, global batch 256, AdamW with learning
rate $10^{-4}$ cosine-decayed to $10^{-5}$, and weight decay $10^{-4}$.
The shared system distributes updates across nine self/cross routes; each directional
Transformer bank distributes its budget across six cross routes. The matched Transformer
uses the same training schedule, data, loss, and seed as the full bank, with widths selected
by parameter count. It has 1,090,074 parameters versus 1,090,081 for the shared system.
Parameter counts exclude body layers.

\subsection{Compact Conversion at Low Latency}
\begin{table}[t]
\centering\footnotesize
\setlength{\tabcolsep}{2pt}
\caption{Input-reference MPJPE (mm), all 9,009 frames and the same six cross routes.
Mean equally weights cohorts. Params count complete converter systems in millions.}
\label{tab:conversion}
\begin{tabular}{lrrrrr}
\toprule
System & Params & Blocked & Motion & Identity & Mean\\
\midrule
Shared & 1.090 & 15.140 & 16.302 & 18.959 & 16.801\\
Transformer & 2.887 & 15.036 & 16.180 & 17.953 & 16.390\\
Matched Transformer & 1.090 & 15.883 & 17.336 & 23.378 & 18.866\\
\bottomrule
\end{tabular}
\end{table}
The 1.090M-parameter shared converter covers nine routes and reaches 16.801 mm
input-reference MPJPE, measuring agreement with input geometry. The 2.887M Transformer
bank reaches 16.390 mm with 2.65 times as many parameters. Compressing that bank to
1.090M increases its mean by 2.476 mm: 0.847 mm on blocked frames, 1.156 mm on new
motions, and 5.425 mm on held-out identities. At matched capacity, the shared converter
reduces mean error by 2.065 mm and identity-cohort error by 4.419 mm. It therefore
approaches the larger bank's accuracy while preserving more accuracy at the matched
parameter budget, especially on the two held-out identities.

\begin{table}[t]
\centering\small
\caption{Batch-one latency on RTX 4090. Official tools cache identity state.
Shared values are median/P95 ms; speedup compares medians.}
\label{tab:latency}
\begin{tabular}{lrrr}
\toprule
Route & Official ms & Shared ms & Speedup\\
\midrule
MHR$\to$SOMA & 49.364 & 7.669 / 8.315 & 6.44$\times$\\
SMPL-X$\to$SOMA & 48.223 & 7.434 / 7.846 & 6.49$\times$\\
SOMA$\to$SMPL-X & 41.830 & 7.486 / 7.810 & 5.59$\times$\\
\bottomrule
\end{tabular}
\end{table}
Timing uses one RTX 4090, four CPU threads, CUDA synchronization, and three calls
on each of nine predetermined frames. The path includes source FK/centering,
normalization, conversion, and target-joint FK; it excludes loading, I/O, world Kabsch,
communication, and robot execution. Cached official MHR/SOMA tools retain pose
conversion~\cite{ferguson2025mhr,saito2026soma}. The three same-GPU directions take
7.43--7.67 ms, 5.59--6.49$\times$ faster than those compositions
(Table~\ref{tab:latency}); measured P95 stays below 8.32 ms. On a separate, matched
300-frame/five-route accuracy subset, official/shared/Transformer means are
13.014/17.266/16.884 mm. This joint accuracy/runtime comparison describes
the operating trade-off; the 300-frame and 9,009-frame scopes remain separate.

\subsection{Extension and Correspondence}
Freezing all original modules preserves their computation by construction. The ANNY
extension therefore evaluates whether a new native interface can be learned around
the fixed representation.
After 945k updates on six new cross routes and ANNY self-reconstruction, new cross
routes reach 22.950 mm input-reference MPJPE, spanning 18.221--29.248 mm across
directions. The complete 16-route system gives 16.723 mm; original-route errors
reproduce the parent within 0.001 mm. Section~\ref{sec:policy-compatibility} then tests
whether the new encoder's features are also interpretable by an existing policy.

The selected representation gives same-frame/wrong-frame cosine 0.982/0.845 on
normalized decoder-input tensors. Removing paired terms throughout gives
0.333/0.011. Centering the selected features retains same-frame cosine 0.920,
distinguishing frame-varying correspondence from their shared static component.
Two branches sharing a 190k ancestor and $10^{-3}$ diversity test retention:
the branch that removes paired losses by 240k still reaches 0.956 same-frame cosine
at 945k. Early correspondence therefore persists during continued reconstruction. These feature
measurements motivate testing consistency in the downstream action space as well
as accuracy in joint space.

\section{Shared Humanoid Control}
\label{sec:control}
\subsection{Frozen Slots as Motion Commands}
\begin{figure*}[t]
\centering
\includegraphics[width=\textwidth]{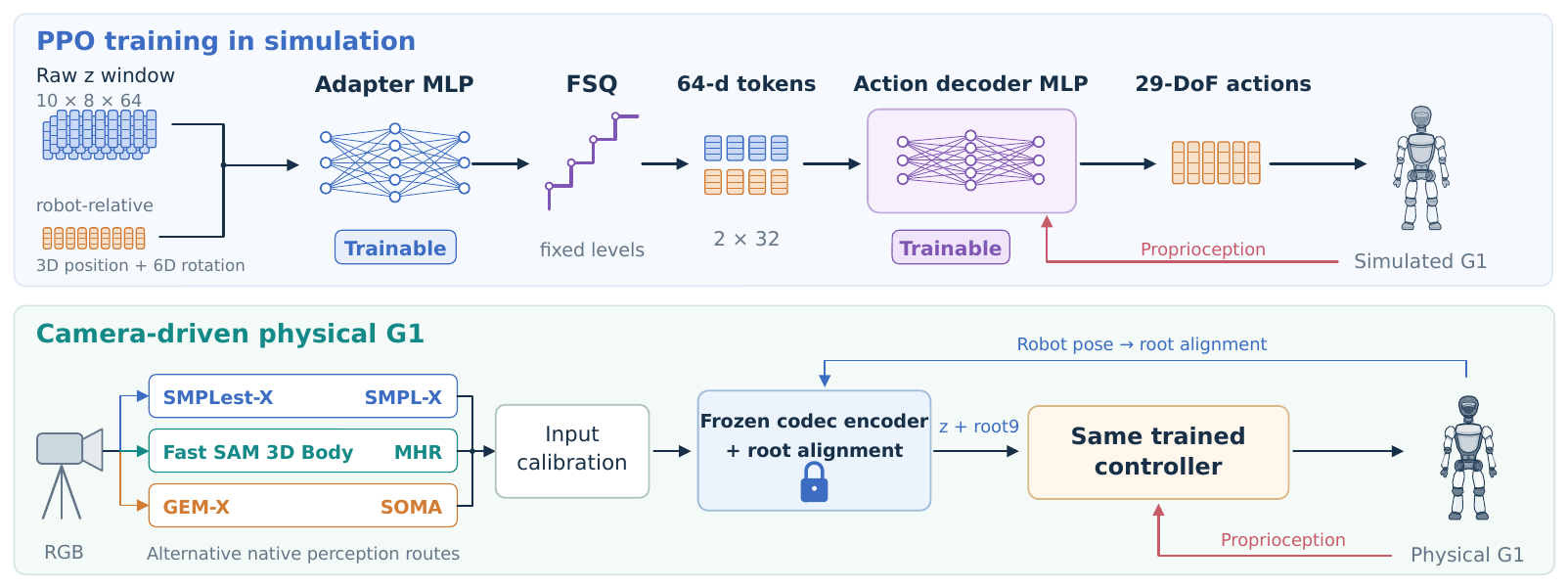}
\caption{Shared control. Top: ten frames of raw slots and robot-relative root
commands feed an adapter and FSQ; the action decoder combines control tokens with
proprioceptive history. PPO trains the control networks with the body codec frozen.
Bottom: three calibrated visual formats share the same simulation-trained G1
controller, with all learned weights fixed during deployment.}
\label{fig:control-pipeline}
\end{figure*}
The controller reads raw $8\times64$ slots, bypassing human-body output heads
(Fig.~\ref{fig:control-pipeline}). Each ten-frame window concatenates 512 latent
channels and nine source-root channels per frame (3D position, 6D rotation), forming
5,210 channels. Source-only rigid placement and measured robot pose express the
root command in robot-relative coordinates. A LayerNorm/SiLU MLP
($5210\to512\to256\to64$) produces two 32-dimensional tokens. Fixed finite scalar
quantization (FSQ), 32 levels per component, passes gradients with a straight-through
estimator. Tokens join 930 proprioceptive-history channels for the 29-action G1 decoder.
Actor and critic have hidden widths $4096,4096,2048,2048,1024,1024,512,512$.

The adapter, action decoder, action distribution, and critic train from random
initialization using a SONIC-style PPO interface~\cite{luo2026sonic}; codec and native
normalizers are frozen. Eight ranks, each with 336 Isaac Lab environments and a
24-step rollout, produce 64,512 transitions/update. PPO uses five epochs, eight
global minibatches, clip 0.2, discount 0.99, and GAE 0.95. Retargeted G1 references
supply tracking rewards and failure checks. No teacher actions or pretrained SONIC
policy weights initialize these learned controllers.

\subsection{One Policy across Three Formats}
\begin{table}[t]
\centering\small
\caption{Three-format control on 1,203 windows and 1,053 complete format/action cases.
The first three policies use a matched PPO budget. No-paired slots omit
$\mathcal L_{\rm cos}$ and $\mathcal L_{\rm match}$ during representation training.
Released SONIC is a separate pretrained reference with its native input interface.}
\label{tab:control}
\begin{tabular}{lrr}
\toprule
Policy input & Window success & Full-action success\\
\midrule
Aligned slots & 79.55\% (957) & 60.78\% (640)\\
No-paired slots & 77.22\% (929) & 56.98\% (600)\\
Joint coordinates & 76.31\% (918) & 57.45\% (605)\\
\midrule
Released SONIC & 80.13\% (964) & 73.98\% (779)\\
\bottomrule
\end{tabular}
\end{table}
This experiment evaluates representation choice for shared-policy learning: MHR,
SMPL-X, and SOMA slots enter one policy. Trained policies use matching random
initial weights, G1 targets, rewards, root caches, and 23,429 PPO updates
(1,511,451,648 transitions), with seed zero. The joint-coordinate baseline pads 72 SMPL24
joint-coordinate channels to 512 and uses the matching adapter. Validation fixes
401 paired windows and 351 paired full actions, each in all three formats. Success
requires completion of every reference frame without failure termination.

With 23,429 PPO updates, the aligned-slot policy reaches 60.78\% full-action success,
exceeding the joint-coordinate baseline (57.45\%) by 3.33 percentage points and no-paired slots
(56.98\%) by 3.80 points (Table~\ref{tab:control}). One policy applies the aligned representation across
MHR, SMPL-X, and SOMA, showing its utility for training a shared controller from
scratch. Released SONIC reaches 73.98\% with its pretrained checkpoint and native
input interface, and serves as a separate reference. Per-format success is 62.68\%
MHR, 60.11\% SMPL-X, and 59.54\% SOMA. On the 145 full actions completed by both
slot policies in all formats, pelvis-relative tracking MPJPE is 33.06/33.67 mm for
aligned/no-paired slots.

\begin{figure*}[t]
\centering
\includegraphics[width=0.88\textwidth]{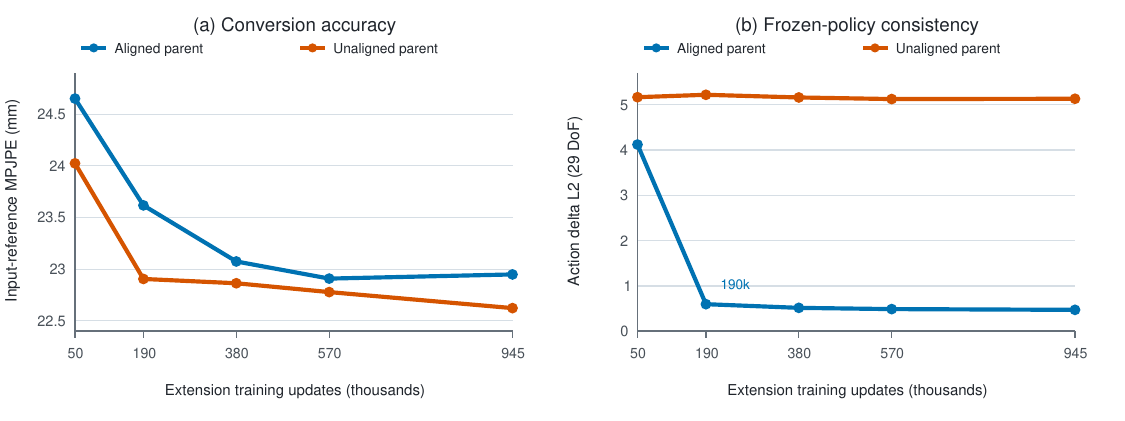}
\caption{Geometric accuracy and frozen-policy compatibility during ANNY extension.
(a) Six-route input-reference MPJPE. (b) Static-pose action deviation on 9,009 frames
from 74 motions: each pose fills ten input positions, with robot state fixed.
Action deviation uses the aligned SMPL-X action as the common reference under the
same frozen controller. Values equally weight motions within each cohort, then the three cohorts.}
\label{fig:anny-policy-consistency}
\end{figure*}

\subsection{Learning from 160 Hours of Motion}
\begin{table}[t]
\centering\small
\caption{Source-weighted full-motion success (\%) on the same 1,240 held-out validation
motions. The frozen-slot policy uses 83,127 PPO updates; released SONIC uses its
pretrained checkpoint and native input interface.}
\label{tab:scaling}
\begin{tabular}{lrr}
\toprule
Controller & PPO updates & Validation\\
\midrule
Frozen slots & 83,127 & 66.10\\
Released SONIC & --- & 62.59\\
\bottomrule
\end{tabular}
\end{table}
To test reuse with a larger motion corpus, a separately initialized controller uses
the same frozen representation and control architecture with 160 hours of motion: 80 hours from
BONES-SEED~\cite{bonesseed2026} and 80 hours from our in-house motion data.
Native SMPL-X inputs are paired with 50 Hz retargeted G1 references. Training totals
83,127 PPO updates, with the body codec fixed throughout.

Evaluation uses 1,240 held-out validation motions. Success is averaged
within each source and weighted by its share of training motion duration.
The frozen-slot controller reaches 66.10\% source-weighted validation success
(Table~\ref{tab:scaling}); the released SONIC checkpoint reaches 62.59\% on the same
motions. This result shows that the aligned representation also supports training a
controller from scratch on a larger motion corpus, with competitive validation
performance under the 160-hour and 83,127-update budget.

\subsection{Frozen-Policy Compatibility of a New Interface}
\label{sec:policy-compatibility}
A new interface should preserve how an existing policy interprets motion.
The aligned and unaligned parents are trained independently; the latter omits
paired correspondence losses throughout pretraining. Both ANNY extensions use the
same frozen 83,127-update controller trained on aligned SMPL-X slots. For each
paired pose, the controller action produced from the aligned parent's SMPL-X slots
serves as the common reference. We test all 9,009 validation poses from 74 motions, repeating
each pose's slots across the ten input positions. All inputs use identical zero
root translation, identity root rotation, and upright-idle proprioception. We report
\begin{equation}
 D_\pi=\frac{1}{3}\sum_{c=1}^{3}\frac{1}{|\mathcal M_c|}
 \sum_{m\in\mathcal M_c}\bar\delta_m,
 \label{eq:action-consistency}
\end{equation}
where $\bar\delta_m$ is the within-motion mean of
$\|\mathbf a^{\rm ANNY}-\mathbf a^{\rm SMPL\mbox{-}X}\|_2$ and $\mathcal M_c$ is the
motion set of cohort $c$. Thus, motions within each cohort and the three cohorts
receive equal weight. Each $\mathbf a\in\mathbb R^{29}$ is a dimensionless deterministic
policy output before joint-specific scaling and execution clipping.

Figure~\ref{fig:anny-policy-consistency} separates geometric accuracy from policy
compatibility. At 945k updates, aligned/unaligned-parent ANNY gives 22.950/22.624 mm
conversion MPJPE and 0.472/5.130 action deviation, respectively; the latter is
90.8\% lower for the aligned-parent extension. Thus, similar geometric fidelity can accompany substantially different
responses to a change of input format. Joint-space reconstruction error alone does
not characterize compatibility with a frozen policy; action consistency measures
whether the new interface preserves the controller's interpretation of motion.
Together with shared-policy learning in Table~\ref{tab:control}, this result connects
cross-format correspondence to downstream use. This offline test evaluates adding a
new body-model input interface to an existing humanoid controller without modifying
or retraining the policy.

For aligned-parent ANNY, action deviation falls from 4.119 at 50k updates to 0.597 at
190k, then continues to 0.472 at 945k. Most of the observed reduction is attained by
190k updates on the same training corpus.

\subsection{Camera-Driven Physical Deployment}
\begin{figure}[t]
\centering
\includegraphics[width=\columnwidth]{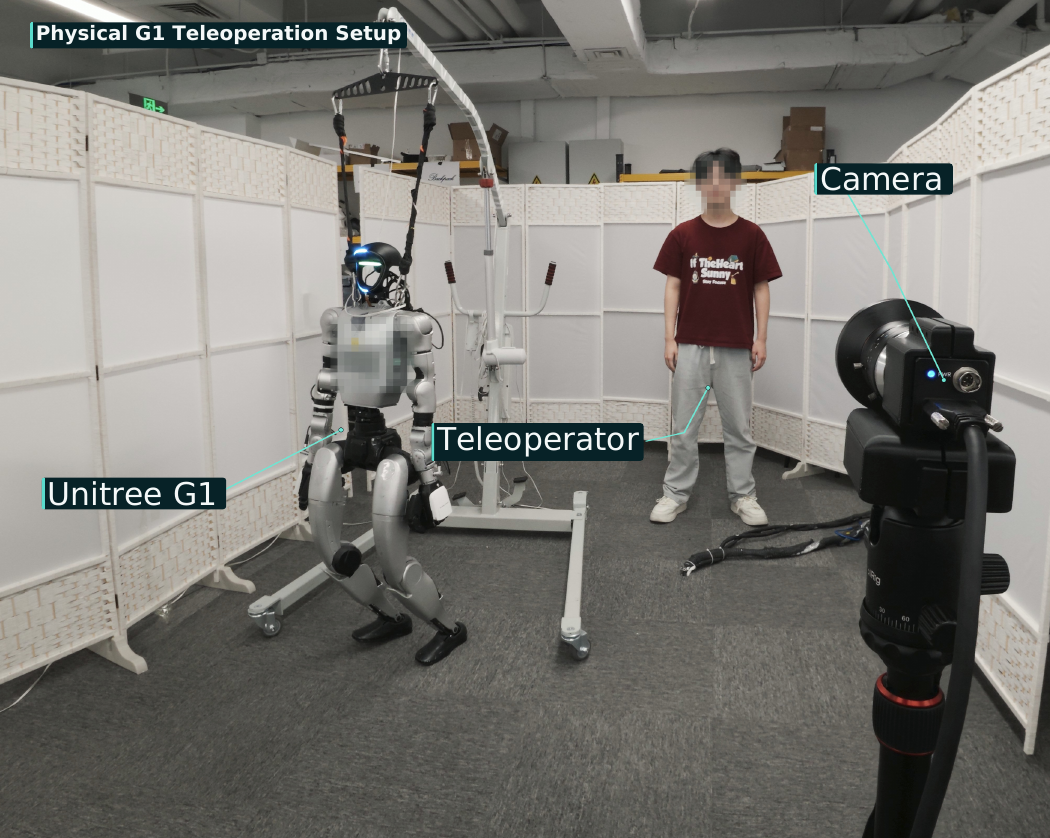}
\caption{Physical G1 teleoperation setup. A camera captures the operator while the Unitree G1
executes motion commands with a fall-arrest line. The three native-format perception
routes and the released SONIC reference use this same setup in sequential trials.}
\label{fig:physical-teleoperation-setup}
\end{figure}

The 160-hour controller drives a physical 29-DoF G1 in the setup shown in
Fig.~\ref{fig:physical-teleoperation-setup}, using three visual front ends:
SMPLest-X$\to$SMPL-X~\cite{yin2026smplestx}, Fast SAM 3D Body$\to$MHR~\cite{yang2026fastsam3dbody},
and GEM-X$\to$SOMA~\cite{nvlabs2026gemx}. Native estimates pass through input
calibration, the frozen codec, and root alignment to the same trained controller.
The robot stands on the floor with a fall-arrest line; target locking and neutral
registration are reset for sequential live trials. Calibration selects valid body
shape, bounds MHR shoulder/elbow corrections, and registers SOMA neutral arm geometry.
It uses observed human geometry rather than robot action targets.

All three routes were exercised in live following with the learned weights fixed.
SMPL-X-derived slots trained this controller, so MHR/SOMA deployment changes the
perception system's native format without changing the codec or policy.
A fourth live reference uses SMPLest-X with the released SONIC checkpoint and its
native motion encoder. The C++ execution stack receives 64-dimensional control
tokens and supplies proprioceptive history. These are qualitative deployment results;
the 50 Hz policy target and 520 ms reference buffer are system settings, not measured
end-to-end latency or task-success statistics.

\section{Conclusion}
\method{} learns a reusable motion interface through native body-model conversion.
A compact shared representation supports native decoding, new ANNY interfaces around
a frozen parent, and humanoid policies trained from scratch. The experiments connect
geometric fidelity to downstream use: comparable conversion errors can accompany
different action consistency under a fixed controller, while aligned slots support
shared-policy learning across formats. Larger-corpus training and qualitative G1
deployment further demonstrate reuse of the frozen representation. Evaluating both
native geometry and policy responses makes this reuse explicit as body models and
control applications change.

\IEEEtriggeratref{3}
\bibliographystyle{IEEEtran}
\bibliography{references}
\end{document}